\documentclass{article}

\usepackage{PRIMEarxiv}

\usepackage[utf8]{inputenc} 
\usepackage[T1]{fontenc}    
\usepackage{hyperref}       
\usepackage{url}            
\usepackage{booktabs}       
\usepackage{amsfonts}       
\usepackage{nicefrac}       
\usepackage{microtype}      
\usepackage{lipsum}
\usepackage{fancyhdr}       
\usepackage{graphicx}       
\graphicspath{{media/}}     
\usepackage{amsmath}
\usepackage{amssymb}
\usepackage{booktabs}
\usepackage{float}
\usepackage{caption}
\usepackage{algorithm}
\usepackage{algorithmic}
\usepackage{multirow}
\usepackage{colortbl}
\usepackage[table]{xcolor} 
\usepackage{array}
\usepackage{tabularx}
\usepackage{adjustbox}
\usepackage{siunitx}
\usepackage{cite}

\title{IPv2: An Improved Image Purification Strategy for Real-World Ultra-Low-Dose Lung CT Denoising}

\author{
  Guoliang Gong$^{1}$, Man Yu$^{2}$$^{*}$ \\
  Tianjin University of Science and Technology \\
  \texttt{onekey029@gmail.com} \\
   \And
  Xianghong Meng$^{3}$, Xiaoliang Wang$^{3}$, Zhongwei Zhang$^{3}$ \\
  Department of Radiology, Tianjin Hospital\\
   \\
}

\begin{document}
\maketitle

\begin{abstract}
The image purification strategy constructs an intermediate distribution with aligned anatomical structures, which effectively corrects the spatial misalignment between real-world ultra-low-dose CT and normal-dose CT images and significantly enhances the structural preservation ability of denoising models. However, this strategy exhibits two inherent limitations. First, it suppresses noise only in the chest wall and bone regions while leaving the image background untreated. Second, it lacks a dedicated mechanism for denoising the lung parenchyma. To address these issues, we systematically redesign the original image purification strategy and propose an improved version termed IPv2. The proposed strategy introduces three core modules, namely Remove Background, Add noise, and Remove noise. These modules endow the model with denoising capability in both background and lung tissue regions during training data construction and provide a more reasonable evaluation protocol through refined label construction at the testing stage. Extensive experiments on our previously established real-world patient lung CT dataset acquired at 2\% radiation dose demonstrate that IPv2 consistently improves background suppression and lung parenchyma restoration across multiple mainstream denoising models. The code is publicly available at\url{https://github.com/MonkeyDadLufy/Image-Purification-Strategy-v2}.
\end{abstract}

\section{Introduction}
\label{sec:Introduction}

Computed tomography plays an essential role in the screening and diagnosis of pulmonary diseases\cite{important1}. However, the ionizing radiation associated with CT examinations poses non-negligible health risks, particularly for high-risk populations who require long-term follow-up\cite{important2}. In accordance with the as low as reasonably achievable principle of radiation protection\cite{alara1,alara2}, reducing the radiation dose per scan to the greatest extent possible while preserving diagnostic information becomes a central research problem in medical imaging. Low-dose CT reduces radiation exposure by decreasing the tube current or tube voltage. Nevertheless, when the dose is reduced to 20\% of the conventional level or even lower\cite{20reduced}, the signal-to-noise ratio deteriorates sharply. The intensity and complexity of noise and artifacts substantially exceed those observed in standard low-dose settings, which poses severe challenges to image denoising algorithms.

In recent years, the rapid development of deep learning has opened a new technical path for low-dose CT image denoising. However, existing studies generally suffer from two paradigm-level limitations. First, research efforts are largely concentrated on achieving marginal performance gains on synthetic datasets. Public benchmarks such as MAYO2016 simulate low-dose images by adding Poisson noise to the raw projection data of normal-dose scans\cite{mayo2016,mayo2020}. Although such synthetic data facilitate algorithm validation and quantitative comparison, they exhibit an intrinsic distribution gap from real clinical scans, in which noise is strongly coupled with anatomical structures\cite{piglet,phantom_dataset}. Second, a limited number of studies based on real patient data attempt to overcome the limitations of synthetic datasets, yet they inevitably face a more challenging issue of structural misalignment\cite{psp,IPDM}. Since ultra-low-dose and normal-dose scans are acquired at different time points, physiological motion leads to noticeable spatial displacement of the same anatomical structures within paired images. Direct end-to-end supervised training on such unaligned data causes the model to learn incorrect pixel-wise mappings, which results in blurred tissue boundaries and distorted anatomical structures in the denoised outputs, thereby compromising diagnostic reliability.

To address the above challenges, our previous work proposes an image purification strategy\cite{ipv1}. This strategy transfers the anatomical contours of normal-dose images to ultra-low-dose images through edge binarization, common mask computation, and orthogonal decomposition of residual vectors. It constructs an intermediate distribution that is structurally aligned with normal-dose images as training data, thereby effectively correcting the spatial misalignment between real-world ultra-low-dose CT and normal-dose CT image pairs. On this basis, the FFM(frequency-domain flow matching) model achieves leading performance in anatomical structure preservation due to its ability to decouple frequency-domain features\cite{ipv1}.

Despite these advances, the image purification strategy still exhibits two unresolved limitations. First, the strategy focuses exclusively on noise suppression in the chest wall and skeletal regions, while leaving the background unprocessed. Since the training inputs, namely IPv1(uLDCT) and NDCT, do not contain background noise, the model learns to ignore background regions during training, which results in substantial residual noise in the background of the denoised outputs. Second, the strategy lacks a targeted design for the lung parenchyma. Because the grayscale contrast between lung tissue and surrounding soft tissue and vascular structures is relatively high, the original design assumes that no additional denoising is required in this region. However, experimental results show that under the extreme noise level of 2\% radiation dose, the fine textures within the lung parenchyma are severely overwhelmed by noise, and the denoising performance in this region becomes nearly ineffective. These two limitations significantly restrict the applicability of the image purification strategy in full-lung clinical diagnosis scenarios.

To overcome the above limitations, this paper systematically reconstructs the original image purification strategy and proposes a new strategy termed IPv2. Building upon the original design, IPv2 introduces three core modules with explicit functionality and clear motivation. The first module, \textbf{Remove Background}, eliminates irrelevant background pixels during common mask computation through a flood-filling algorithm\cite{floodfill}. As a result, the synthesized training data preserve the background noise distribution of real ultra-low-dose images, which guides the model to learn background denoising capability. The second module, \textbf{Add Noise}, performs the Radon transform on normal-dose images\cite{radontransform}, adds Poisson-Gaussian mixed noise, and then applies the inverse Radon transform. This process injects synthetic noise into the lung parenchyma with intensity matched to real ultra-low-dose noise, so that the simulated ultra-low-dose inputs used in training exhibit realistic noise characteristics in lung tissue regions, thereby enabling the model to learn targeted denoising for the lung parenchyma. The third module, \textbf{Remove Noise}, trains a weak denoiser on simulated ultra-low-dose images. During test label construction, the original ultra-low-dose image is first processed by this weak denoiser to obtain an intermediate image in which noise within the lung tissue is suppressed. The lung parenchyma region from this intermediate image is then fused with the chest wall, skeletal, and background regions from the normal-dose image. The resulting label image is fully noise-free in the lung parenchyma, which provides a more reliable reference for model evaluation. These three modules operate at the stages of training data construction and test label construction, respectively, and together constitute the complete IPv2 strategy.

\begin{figure*}[t]
    \centering
    \includegraphics[width=\textwidth]{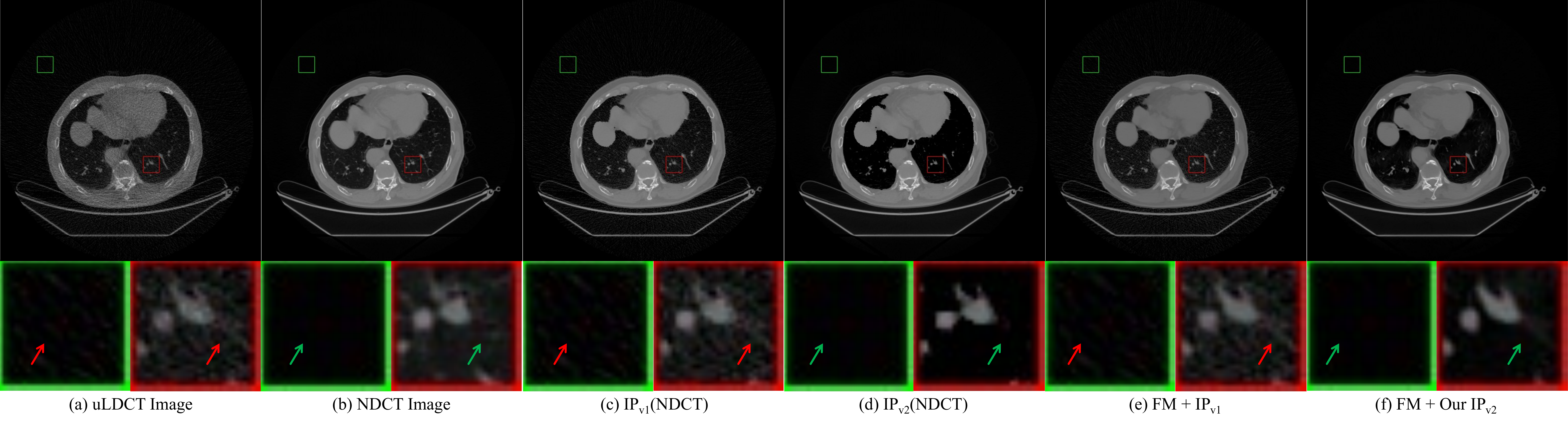}
    \captionsetup{width=\linewidth} 
    \caption{Motivation of the improved image purification strategy (IPv2). (a) Ultra-low-dose CT (uLDCT) image with severe noise that degrades structural clarity. (b) Corresponding normal-dose CT (NDCT) image. (c) NDCT processed by the IPv1 strategy, which serves as the test label. (d) NDCT processed by the IPv2 strategy. (e) Denoising result of (a) obtained by training a flow matching\cite{FM} (FM) model in an end-to-end manner under the IPv1 strategy. (f) Denoising result of (a) obtained by training a FM\cite{FM} model in an end-to-end manner under the IPv2 strategy. The \textcolor{green}{green} enlarged boxes in (a) to (f) indicate background regions, and the \textcolor{red}{red} enlarged boxes indicate lung tissue regions. The \textcolor{red}{red} arrows denote the presence of noise, while the \textcolor{green}{green} arrows denote the absence of noise. Please zoom in for details.}
    \label{fig:1}
\end{figure*}

Figure~\ref{fig:1}(a) shows an ultra-low-dose CT image with severe noise contamination. Figure~\ref{fig:1}(b) presents the corresponding normal-dose CT image. A comparison of the red enlarged regions in Figure~\ref{fig:1}(a) and (b) reveals evident discrepancies in anatomical morphology. Training a network directly on such unaligned image pairs causes the model to learn incorrect mapping relationships, which leads to anatomical distortion in the denoised results. In particular, the original tissue structures in the input uLDCT image cannot be faithfully preserved, and such structural inconsistency directly undermines diagnostic accuracy.

Figure~\ref{fig:1}(c) illustrates the NDCT processed by the original image purification strategy, which serves as the evaluation label for denoising networks, while Figure~\ref{fig:1}(d) shows the NDCT processed by the improved purification strategy. A comparison between these two labels indicates that, in the background and lung tissue regions, the label generated by IPv2 contains substantially less residual noise than that produced by IPv1.

Figure~\ref{fig:1}(e) and (f) further compare the denoising performance before and after the proposed improvement. Figure~\ref{fig:1}(e) presents the result obtained by training a FM\cite{FM} model in an end-to-end manner on the data shown in Figure~\ref{fig:1}(a). The red arrows indicate that the model fails to suppress noise in the background and lung tissue regions. In contrast, Figure~\ref{fig:1}(f) shows the denoised output of the FM\cite{FM} model trained under the IPv2 strategy. The green arrows demonstrate that, while maintaining the denoising capability of IPv1 in the chest wall and skeletal regions, IPv2 further enhances noise suppression in the background and lung tissue regions.

In summary, the main contributions of this work are as follows. 

\begin{itemize}
    \item  We propose an improved image purification strategy termed IPv2, which introduces three core modules, namely \textbf{Remove Background}, \textbf{Add Noise}, and \textbf{Remove Noise}, and systematically addresses the fundamental limitation of the original strategy in handling background and lung tissue denoising.
    \item We conduct extensive and systematic experiments on the previously established real patient lung CT dataset acquired at 2\% radiation dose. The results demonstrate that IPv2 consistently improves the performance of various mainstream denoising models and reveal the fundamental constraint imposed by training data quality on ultra-low-dose CT denoising.
\end{itemize}

\section{RELATED WORK}
\label{sec:RELATED WORK}

\subsection{Methods for Addressing Data Misalignment}
\label{sub:解决数据不对齐的方法}

Image registration is a classical solution to spatial misalignment in medical image analysis\cite{image_registration_survey}. Conventional approaches include rigid\cite{Rigid_registration}, affine\cite{affine_registration}, and non-rigid deformable registration\cite{non-rigid_deformation_registration}. In recent years, deep learning based models such as VoxelMorph and uniGradICON surpass traditional methods in both registration accuracy and inference speed\cite{VoxelMorph,unigradicon}. However, these approaches encounter dual challenges in the ultra-low-dose CT scenario. On the one hand, extreme noise severely disrupts stable extraction of anatomical boundaries. On the other hand, large deformations caused by respiratory motion impose stringent requirements on the smoothness and topology preservation of the deformation field. Empirical results show that even uniGradICON, which is designed with zero-shot generalization capability, still produces registration outcomes with global resolution degradation and local structural misalignment when applied to ultra-low-dose and normal-dose image pairs.

Data cleaning strategies, represented by the PSP method\cite{psp}, attempt to mitigate misalignment by cropping images into local patches and computing the similarity of binary masks between paired patches. Only patch pairs with highly consistent anatomical structures are retained as training samples. This strategy performs well in conventional low-dose CT denoising tasks. However, when applied to the ultra-low-dose setting at 2\% radiation dose, the extremely low signal-to-noise ratio renders most local patches structurally indistinguishable and thus classified as dissimilar. Consequently, the training dataset shrinks drastically, and the model suffers from underfitting, which prevents recovery of diagnostically meaningful details.

The image purification strategy is first proposed in our previous work\cite{ipv1}. It extracts anatomical contours from ultra-low-dose and normal-dose images through edge binarization, computes a common mask as a structural weighting term, and orthogonally decomposes the residual vector from normal-dose to ultra-low-dose images into structural and texture variation components. Based on this decomposition, it constructs purified images that are anatomically aligned with normal-dose images while preserving ultra-low-dose texture characteristics, together with their corresponding normal-dose images as training pairs. This strategy effectively resolves structural misalignment without reducing the original dataset size and establishes a high-quality training foundation for ultra-low-dose CT denoising. Nevertheless, its original design focuses primarily on noise suppression in the chest wall and skeletal regions, without dedicated treatment of the background and lung parenchyma. The IPv2 strategy proposed in this work constitutes a systematic refinement to address these limitations.

\subsection{Low-Dose CT Image Denoising Networks}
\label{sub:低剂量CT图像去噪网络}

Direct mapping approaches are represented by supervised non-diffusion models that perform end-to-end transformation from noisy to clean images through a single forward pass. REDCNN achieves baseline performance with an encoder-decoder residual architecture\cite{redcnn}. EDCNN introduces an edge enhancement module to improve anatomical contour fidelity\cite{edcnn}. DU-GAN employs dual discriminators in the image and gradient domains to jointly constrain the generation process\cite{DU-GAN}. MMCA combines a vision Transformer with the Sobel operator to strengthen edge feature extraction\cite{vit,MMCA}. These methods offer fast inference and stable training. However, in the ultra-low-dose setting, their capability to reconstruct fine texture details is substantially limited. More critically, when structural misalignment exists in the training data, these models tend to learn incorrect mapping relationships, which leads to anatomical distortion in the denoised outputs.

Iterative mapping approaches are centered on diffusion-based models that progressively recover image details through multiple refinement steps. Cold Diffusion generalizes the noise addition mechanism of conventional diffusion models to arbitrary image transformations\cite{coldDiffusion}. CoreDiff introduces an error-regulated restoration network to correct sampling bias\cite{corediff}. Flow Matching defines conditional probability paths based on optimal transport displacement interpolation to enable efficient sampling\cite{FM,Optimal_Transport}. These methods demonstrate clear advantages in texture reconstruction, yet their performance heavily depends on structural consistency in the training data. The theory of Rectified Flow indicates that when the mapping path between data distributions contains numerous intersections\cite{reflow}, the difficulty and uncertainty of sampling increase significantly. We further argue that structural misalignment between ultra-low-dose and normal-dose images constitutes a primary cause of such path intersections. Therefore, improving structural alignment in training data is a prerequisite for fully exploiting the potential of iterative mapping models.

Prior-guided mapping approaches, typically based on self-supervised learning, rely on noise statistics in the absence of paired data. N2N constructs training pairs by repeatedly adding noise to the same low-dose image\cite{N2N}. IPDM uses only normal-dose images and generates approximate low-dose images by injecting noise into early diffusion states\cite{IPDM}. Although these methods avoid the misalignment issue, they assume that noise follows a known distribution and is independent of the underlying signal. This assumption deviates substantially from real ultra-low-dose CT images, in which noise is strongly coupled with anatomical structures, resulting in limited denoising performance on real clinical data.

Neither image registration, data cleaning strategies, nor various denoising network architectures alone resolve the dual challenge of structural misalignment and complex noise suppression in real-world ultra-low-dose CT denoising. The image purification strategy provides an effective solution at the data level. However, its insufficient denoising capability in the background and lung parenchyma regions remains the primary bottleneck. Addressing this limitation constitutes the central objective of the proposed IPv2 strategy.

\section{PROPOSED METHOD}
\label{sec:PROPOSED METHOD}

\begin{figure*}[htbp]
    \centering
    \includegraphics[width=\textwidth]{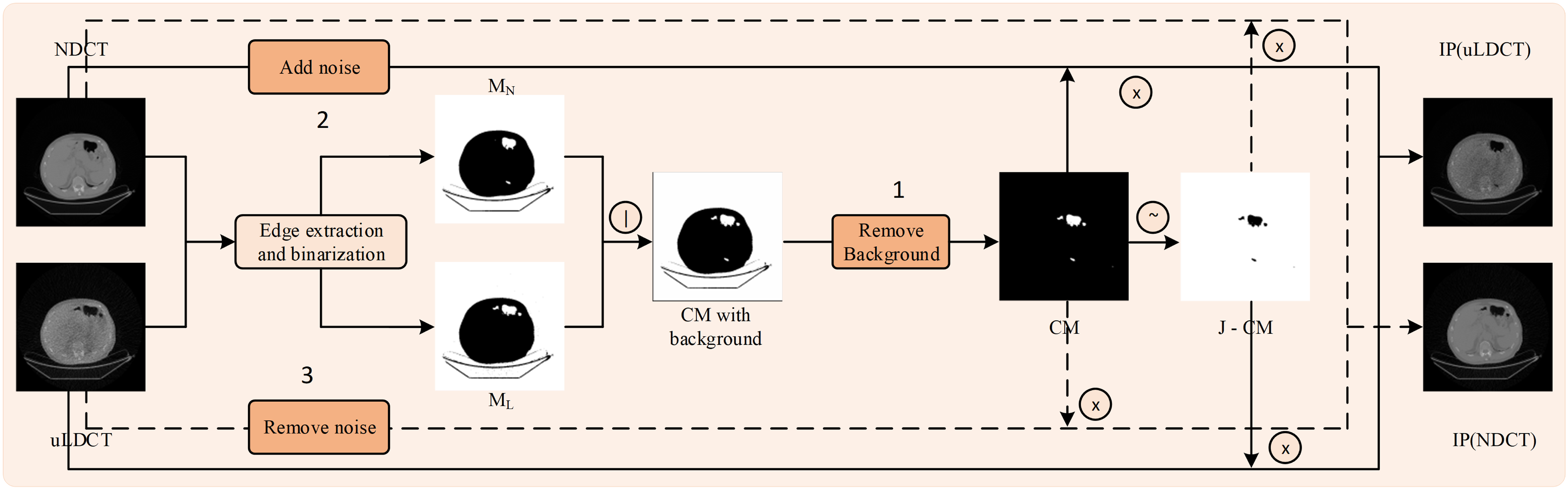}
    \captionsetup{width=\linewidth} 
    \caption{Overview of the proposed improved image purification strategy (IPv2).}
    \label{fig:2}
\end{figure*}

To address the limited denoising capability of the original image purification strategy in both the background and lung parenchyma regions, we propose an improved image purification strategy termed IPv2. The proposed method preserves the core principle of structural alignment and texture decoupling while introducing three dedicated modules with clear functional roles and explicit motivations. The overall pipeline is illustrated in Figure~\ref{fig:2}.
\subsection{Remove Background Module}
\label{sub:Remove Background模块}

In the original image purification strategy, the common mask is obtained by directly performing a logical OR operation between the binarized masks of the ultra-low-dose and normal-dose images. This design leads to a subtle but critical consequence. Since background pixels take the value 0 in both masks, the logical OR result remains 0, and the background is therefore treated as a region without structural information in the common mask. During subsequent residual decomposition and image synthesis, these pixels are consistently assigned zero weight. As a result, the purified images used for training, denoted as IPv1(uLDCT), contain no background noise, whereas the ultra-low-dose images at test time exhibit substantial background noise. Under strict supervision, the model thus learns to ignore background noise rather than to remove it.

To address this issue, we introduce a flood fill based background removal step before computing the common mask\cite{floodfill}. Specifically, we first apply Otsu adaptive thresholding to the original ultra-low-dose image to obtain an initial binary mask\cite{otsu1975threshold}. We then perform flood fill using the four image corners as seed points and label all regions connected to these seeds as background by setting their mask values to 0. After this process, regions corresponding to anatomical structures such as the chest wall, bones, and lung parenchyma are preserved, while pure background regions are completely removed. The refined mask is then combined with the binarized mask of the normal-dose image using a logical OR operation. In the resulting common mask, background pixels are correctly assigned a weight of 0 rather than being implicitly treated as informative regions.

During image synthesis, we retain the weighted fusion formulation defined in the original strategy. For each pixel location, if it belongs to an anatomical structure, the purified image inherits the pixel value from the normal-dose image after the Add Noise module described in Section~\ref{sub:Add noise模块}. If it belongs to the background, the purified image directly inherits the pixel value from the ultra-low-dose image. The resulting IPv2(uLDCT) image exhibits two key properties. In anatomical regions, it preserves the structural contours of the normal-dose image together with simulated noise texture, while in background regions it maintains the same noise distribution as the input ultra-low-dose image. When trained with such data, the model is compelled to learn a mapping from noisy background to clean background, instead of a trivial identity mapping between clean backgrounds. Consequently, the model acquires effective background denoising capability at inference time.

\subsection{Add Noise Module}
\label{sub:Add noise模块}

The original image purification strategy handles the lung parenchyma under an implicit assumption that the gray level contrast between lung tissue and surrounding structures remains sufficiently high so that visual discrimination is preserved even in the presence of substantial noise. Under this assumption, no additional denoising constraint is imposed on this region. However, when the radiation dose is reduced to 2\% of the normal level, this assumption no longer holds. Empirical observations indicate that in real ultra-low-dose CT images, the fine textures within the lung parenchyma are almost entirely submerged by noise, and the model exhibits nearly ineffective denoising performance in this region. More critically, the simulated ultra-low-dose images used during training, denoted as IPv1(uLDCT), contain no noise in the lung parenchyma, which deprives the model of effective supervision for learning lung-specific denoising.

To equip the model with denoising capability in the lung parenchyma, we introduce an Add Noise module during the construction of the training set. The objective of this module is not to perfectly replicate the true noise distribution of real ultra-low-dose data, but to inject synthetic noise with controllable intensity and approximately matched distribution into the lung region, thereby providing the model with explicit denoising experience during training.

The implementation proceeds as follows. We first apply the Radon transform to the normal-dose image to project it from the pixel domain to the sinogram domain\cite{radontransform}. In the sinogram domain, we sequentially add Poisson noise to model photon counting fluctuations and Gaussian noise to model electronic thermal noise. The noise intensity is calibrated according to the physical relationship between radiation dose and noise level, ensuring that the effective radiation dose of the noised data is consistent with that of the real ultra-low-dose dataset used in this study. We then perform the inverse Radon transform to reconstruct a simulated ultra-low-dose image in the pixel domain.

After this process, the resulting IPv2(uLDCT) image exhibits synthetic noise in the lung parenchyma with intensity matched to that of real ultra-low-dose images, while the chest wall, bones, and background retain the original texture characteristics of the real ultra-low-dose data. By pairing this image with its corresponding normal-dose image for supervised training, the model learns a mapping from noisy lung tissue to clean lung tissue under explicit supervision. Ablation studies show that this module alone yields a substantial improvement in denoising performance within the lung parenchyma.

\subsection{Remove Noise Module}
\label{sub:Remove noise模块}

During testing, the label image used to evaluate model performance, denoted as IPv2(NDCT), should be completely noise-free in the lung parenchyma so that it faithfully reflects the true denoising capability of the model in this region. The Add Noise module operates on training inputs rather than on test labels, and therefore cannot directly address the noise suppression requirement of the evaluation labels.

To resolve this issue, we introduce a Remove Noise module. The core idea exploits a specific property of a denoiser trained on simulated data. Such a denoiser exhibits limited ability to suppress the complex noise patterns in the chest wall and bone regions of real ultra-low-dose images, yet it effectively removes noise in the lung parenchyma. This behavior arises because noise in the chest wall and bone regions presents more complicated structural characteristics that are difficult to reproduce through simulation, whereas noise in the lung parenchyma more closely follows an additive assumption and thus generalizes better from simulated training data.

The implementation consists of two stages. In the training stage, we use the simulated ultra-low-dose images constructed in Section~\ref{sub:Add noise模块} together with the original normal-dose images to train a denoiser. For simplicity, this auxiliary denoiser adopts the same network architecture and parameter scale as the main denoising network, and it is optimized using a standard pixel-wise reconstruction loss. Since the training data contain simulated noise only, the learned model acquires effective denoising capability for the simulated noise distribution. When applied to real ultra-low-dose images, it shows weak suppression of noise in the chest wall and bone regions, while it achieves effective denoising in the lung parenchyma. For this reason, we refer to it as a weak denoiser, whose effectiveness is spatially confined to the lung region and nearly negligible elsewhere.

In the label construction stage, the original ultra-low-dose images in the training set are processed by the weak denoiser to obtain intermediate images in which noise in the lung parenchyma is suppressed, while the chest wall, bone, and background regions retain their original noise characteristics. We then perform image fusion guided by the common mask computed in Section~\ref{sub:Remove Background模块}. For pixels belonging to the chest wall, bone, or background regions, the values are inherited from the original normal-dose image. For pixels belonging to the lung parenchyma, the values are taken from the weakly denoised ultra-low-dose image. The resulting IPv2(NDCT) label image possesses three properties. It preserves the clear texture of the normal-dose image in the chest wall and bone regions, maintains the same noise-free characteristics as the normal-dose image in the background, and achieves complete noise removal in the lung parenchyma.

The design of this module is motivated by two considerations. First, the weak denoiser is used only for constructing evaluation labels and does not participate in the training or inference of the main denoising network, so its spatial limitation does not introduce adverse effects. Second, by combining the weak denoiser with mask-guided image fusion, we achieve precise noise removal in the lung region without compromising the authenticity of texture characteristics in the chest wall, bone, and background regions, thereby providing labels that are closer to an ideal reference for performance evaluation. Ablation studies demonstrate that the joint use of this module with the Remove Background and Add Noise modules yields the best denoising performance.

\subsection{Training and Evaluation Paradigm}
\label{sub:训练与评估范式}

Based on the three modules described above, we establish a complete IPv2 training and evaluation paradigm. During training, for each pair of original ultra-low-dose and normal-dose images in the training set, we first construct an IPv2(uLDCT) image through the Remove Background module and the Add Noise module. This image forms a structurally aligned training pair with the corresponding normal-dose image. The main denoising network is then optimized under supervision using this aligned pair. During evaluation, for each pair of original ultra-low-dose and normal-dose images in the test set, we construct an IPv2(NDCT) image through the Remove Background module and the Remove Noise module. The resulting image serves as a label that is structurally aligned with the ultra-low-dose input and texture-consistent with the normal-dose reference, and it is used to assess the denoising performance of the main network.


\section{EXPERIMENTS}
\label{sec:EXPERIMENTS}

\subsection{Our Real-World Datasets}
\label{sub:Our Real-World Datasets}

To evaluate the effectiveness of the proposed IPv2 strategy, we conduct experiments on a real-world ultra-low-dose lung CT dataset constructed in our prior work\cite{ipv1}. The dataset contains 4310 paired ultra-low-dose and normal-dose images. The ultra-low-dose scans use an acquisition setting of 80kV/10mA, while the normal-dose scans use 120kV/250mA. According to the dose-length product estimation, the radiation dose of the ultra-low-dose scans is approximately 2\% of that of the normal-dose scans. To the best of our knowledge, this dataset represents one of the lowest-dose publicly available real clinical paired datasets, which provides an ideal benchmark for assessing the denoising capability of the proposed strategy under extremely high noise conditions.

Following the data partition protocol established in our prior work\cite{ipv1}, we randomly split the dataset into training, validation, and test sets with a ratio of 7:1.5:1.5, resulting in 3017, 646, and 647 image pairs, respectively. All experiments are conducted under this fixed split to ensure a fair comparison with prior work.
\subsection{Experimental Setup}
\label{sub:实验设置}

\subsubsection{Implementation Details}

To ensure fairness and reproducibility, we standardize the training protocol for all compared models. Each model is trained for 200 epochs with an initial learning rate of $10^{-4}$ and a batch size of 2. The number of time steps is set to 10 for CoreDiff\cite{corediff}, FM\cite{FM}, and FFM\cite{ipv1}, and to 50 for Cold Diffusion\cite{coldDiffusion}. All experiments are conducted on an NVIDIA A100 GPU.

\subsubsection{Evaluation Metrics}

Since real-world ultra-low-dose CT images and normal-dose CT images inevitably exhibit anatomical discrepancies, we do not adopt conventional metrics such as PSNR and SSIM, which rely on pixel-wise alignment. Instead, we employ feature-level distribution distance metrics, including FID, KID, and CLIP-FID, for objective evaluation. These metrics compute the distribution similarity between generated images and the target image set in a deep feature space and remain robust to minor spatial misalignment.

\subsection{Comparison with Baseline Methods}
\label{sub:与基线方法的对比实验}

\begin{table*}[htbp]
\centering
\setlength{\tabcolsep}{10pt}
\captionsetup{width=\textwidth}
\caption{Quantitative results (mean values) of different algorithms on the real patient lung dataset. The improvement ratio over the baseline method is indicated in the preceding column.}
\label{table:1}
\begin{tabularx}{\textwidth}{@{}l >{\raggedleft\arraybackslash}X c c c c c c@{}}
\toprule
\textbf{Label} & \textbf{Method} & \multicolumn{2}{r}{\textbf{FID$\downarrow$}} & \multicolumn{2}{r}{\textbf{KID$\times$100$\downarrow$}} & \multicolumn{2}{r}{\textbf{CLIP-FID$\times$100$\downarrow$}} \\
\midrule
\multirow{8}{*}{\cellcolor{white}NDCT}
& FM\cite{FM} + IPv$_1$ & & 81.78 & & 4.72 & & 4.68 \\
& FM\cite{FM} + IPv$_2$ & 52\% & \textbf{39.33} & 79\% & \textbf{1.00} & 47\% & \textbf{2.50} \\
& CoreDiff\cite{corediff} + IPv$_1$ & & 95.87 & & 5.28 & & 7.03 \\
& CoreDiff\cite{corediff} + IPv$_2$ & 17\% & \textbf{79.10} & 34\% & \textbf{3.47} & 34\% & \textbf{4.66} \\
& Cold Diffusion\cite{coldDiffusion} + IPv$_1$ & & 72.55 & & 3.86 & & 3.56 \\
& Cold Diffusion\cite{coldDiffusion} + IPv$_2$ & 54\% & \textbf{33.67} & 81\% & \textbf{0.72} & 57\% & \textbf{1.53} \\
& FFM\cite{ipv1} + IPv$_1$ & & 79.22 & & 4.64 & & 4.35 \\
& FFM\cite{ipv1} + IPv$_2$ & 66\% & \textbf{27.02} & 88\% & \textbf{0.58} & 74\% & \textbf{1.12} \\
\bottomrule
\end{tabularx}
\end{table*}

\subsubsection{Comparison Methods}
\label{subsub:comparison_methods}
To comprehensively evaluate the effectiveness of the IPv2 strategy, we select four representative iterative mapping denoising networks: CoreDiff\cite{corediff}, Cold Diffusion\cite{coldDiffusion}, FM\cite{FM}, and FFM\cite{ipv1}. Each model is trained and tested under both the original image purification strategy (IPv1) and the proposed IPv2 strategy.

\subsubsection{Quantitative Results}
\label{subsub:quantitative_results}
Table~\ref{table:1} presents the quantitative comparison results of different methods on the test set. The IPv2 strategy consistently and significantly improves the performance of all four denoising networks. Taking the FID metric as an example, FM\cite{FM} decreases from 81.78 under IPv1 to 39.33 under IPv2, a reduction of 52\%; Cold Diffusion\cite{coldDiffusion} drops from 72.55 to 33.67, a reduction of 54\%; and FFM\cite{ipv1} achieves the best result among all methods, with an FID as low as 27.02. CoreDiff\cite{corediff} achieves an FID of 95.87 under IPv1, with an improvement margin of 17\%, which is lower than that of the other three models. Further analysis reveals that the noise sampling strategy of CoreDiff\cite{corediff} is designed for conventional low-dose CT scenarios, limiting its feature extraction capability on ultra-low-dose images with 2\% radiation dose. Nevertheless, the IPv2 strategy still brings non-trivial performance gains to it. FFM\cite{ipv1} achieves comprehensive superiority under IPv2, with FID, KID, and CLIP-FID significantly outperforming those of the other compared methods. This result indicates a synergistic gain effect between high-quality training data and the ability to decouple frequency-domain features.

\subsubsection{Visual Quality}
\label{subsub:visual_quality}
\begin{figure*}[htbp]
    \centering
    \includegraphics[width=\textwidth]{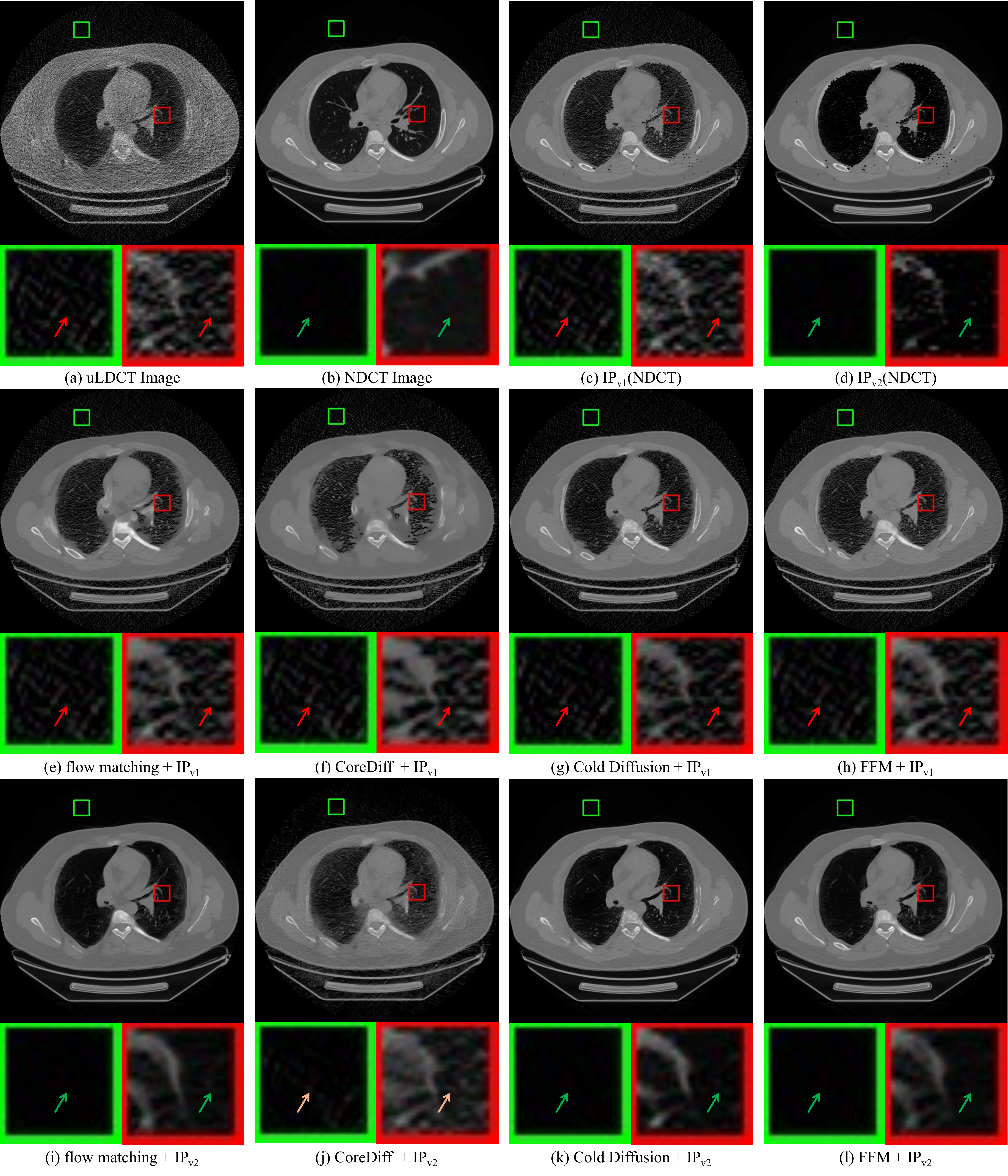}
    \captionsetup{width=\linewidth} 
    \caption{Comparison of visual quality for various denoising methods under the previous purification strategy (IPv1) and the proposed strategy (IPv2). (c) Label constructed with IPv1, (d) Label constructed with IPv2. \textcolor{red}{Red} arrows indicate the presence of noise, \textcolor{green}{green} arrows indicate the absence of noise, and \textcolor{yellow}{yellow} arrows indicate slightly inferior denoising performance.}
    \label{fig:3}
\end{figure*}

Figure~\ref{fig:3} presents a visual quality comparison of different methods on test set samples. Close-up views of the magnified regions clearly show that models trained with the IPv1 strategy fail to achieve effective denoising in both the background and lung parenchyma areas, with noise levels in these regions nearly identical to those in the input ultra-low-dose images. In contrast, models trained with the IPv2 strategy exhibit notable noise suppression in the background, while the textural structures within the lung parenchyma are also markedly restored. This visual comparison aligns closely with the quantitative results discussed above, providing intuitive and compelling evidence for the effectiveness of the IPv2 strategy.

\subsection{Ablation Study}
\label{sub:ablation_study}

\subsubsection{Impact of Modules 1 and 3}
\label{subsub:impact_of_modules_1_and_3}
\begin{figure*}[h]
    \centering
    \includegraphics[width=\textwidth]{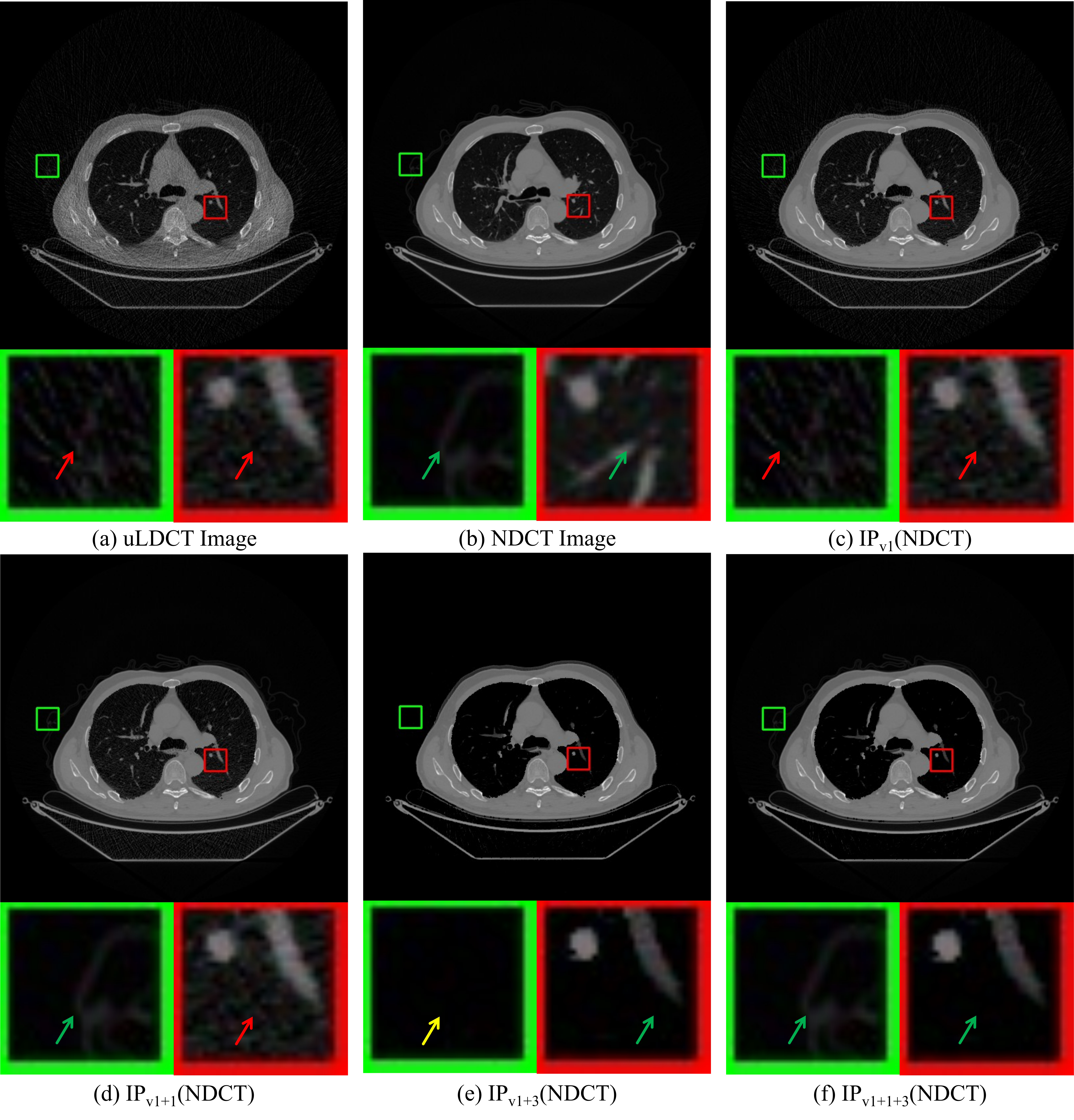}
    \captionsetup{width=\linewidth} 
    \caption{Visualization of processed NDCT results generated by different modules within the purification strategy, which serve as labels during testing. (c) Result from the previous purification strategy (IPv1). (d) to (f) Results from adding Module 1, adding Module 3, and adding both Module 1 and Module 3 to IPv1, respectively. \textcolor{red}{Red} arrows indicate the presence of noise, \textcolor{green}{green} arrows indicate the absence of noise, and \textcolor{yellow}{yellow} arrows indicate the removal of useful information.}
    \label{fig:4}
\end{figure*}

For clarity of presentation, we denote the Remove Background, Add Noise, and Remove Noise modules by the numbers 1, 2, and 3, respectively, as indicated in Figure~\ref{fig:2}. We conduct ablation studies by incorporating Module 1 and Module 3, individually and in combination, into the IPv1 baseline. We then compare the distribution distance between the resulting processed images (IPv2(NDCT)) and the original NDCT. The experimental results, shown in Table~\ref{table:2}, indicate that, compared to IPv1, integrating either Module 1 or Module 3 alone brings the processed images closer to the NDCT distribution. Adding both modules simultaneously further enhances the strategy's capability. Figure~\ref{fig:4} provides an intuitive illustration of the underlying mechanism. Using Module 1 alone effectively denoises the background but fails to denoise the lung tissue regions. Conversely, using Module 3 alone denoises the lung tissue but may remove important information from the background. The combination of Module 1 and Module 3 leverages the strengths of both, achieving effective denoising in both the background and lung tissue areas.

\begin{table}[H]
\centering
\setlength{\tabcolsep}{14pt}
\caption{Results of applying different modules to NDCT in the patient dataset.The improvement ratio over the baseline method is indicated in the preceding column.}
\label{table:2}
\begin{adjustbox}{width=\columnwidth}
\begin{tabularx}{\columnwidth}{@{}l >{\raggedright\arraybackslash}X c c c c c c@{}}
\toprule
\textbf{Label} & \textbf{NDCT} & \multicolumn{2}{c}{\textbf{FID$\downarrow$}} & \multicolumn{2}{c}{\textbf{KID$\times$100$\downarrow$}} & \multicolumn{2}{c}{\textbf{CLIP-FID$\times$100$\downarrow$}} \\
\midrule
\multirow{4}{*}{\cellcolor{white}NDCT}
& IPv$_1$ & & 85.79 & & 5.04 & & 4.68 \\
& IPv$_{1}+1$ & 35\% & \textbf{55.98} & 60\% & \textbf{2.04} & 43\% & \textbf{2.65} \\
& IPv$_{1}+3$ & 39\% & \textbf{52.09} & 70\% & \textbf{1.52} & 23\% & \textbf{3.62} \\
& IPv$_{1}+1+3$ & 58\% & \textbf{36.34} & 82\% & \textbf{0.89} & 47\% & \textbf{2.49} \\
\bottomrule
\end{tabularx}
\end{adjustbox}
\end{table}

\subsubsection{Choice of Label}
\label{subsub:choice_of_label}
IPv2(NDCT) retains the structure of uLDCT while exhibiting the texture of NDCT, making it a more suitable label for evaluating model denoising performance. To assess the impact of the label choice, we consider the improvement in FID, KID, and CLIP-FID metrics rather than their absolute values. As shown in Table~\ref{table:3}, the choice of label has a limited effect on denoising results under FID and KID. However, under CLIP-FID, using IPv2(NDCT) as the label yields a more optimistic evaluation of the FM\cite{FM} denoising result compared to using NDCT (59\% improvement versus 47\%) when the label is constructed with IPv1 combined with Modules 1 and 2. Overall, these findings suggest that IPv2(NDCT) can serve as an effective label for evaluating model denoising performance.

\begin{table}[H]
\centering
\setlength{\tabcolsep}{6pt}
\caption{Denosing results of training FM\cite{FM} on the patient dataset with different labels and modules.The improvement ratio over the baseline method is indicated in the preceding column.}
\label{table:3}
\begin{adjustbox}{width=\columnwidth}
\begin{tabularx}{\columnwidth}{@{}l >{\raggedright\arraybackslash}X c c c c c c@{}}
\toprule
\textbf{Label} & \textbf{Method} & \multicolumn{2}{c}{\textbf{FID$\downarrow$}} & \multicolumn{2}{c}{\textbf{KID$\times$100$\downarrow$}} & \multicolumn{2}{c}{\textbf{CLIP-FID$\times$100$\downarrow$}} \\
\midrule
\multirow{4}{*}{\cellcolor{white}NDCT + IPv$_2$}
& FM\cite{FM} + IPv$_1$ & & 89.27 & & 5.35 & & 4.54 \\
& FM\cite{FM} + IPv$_{1}+1$ & 51\% & \textbf{43.84} & 74\% & \textbf{1.39} & 54\% & \textbf{2.11} \\
& FM\cite{FM} + IPv$_{1}+2$ & 38\% & \textbf{55.23} & 71\% & \textbf{1.57} & 43\% & \textbf{2.58} \\
& FM\cite{FM} + IPv$_{1}+1+2$ & 52\% & \textbf{42.90} & 79\% & \textbf{1.12} & 59\% & \textbf{1.85} \\
\midrule
\multirow{4}{*}{\cellcolor{white}NDCT}
& FM\cite{FM} + IPv$_1$ & & 81.78 & & 4.72 & & 4.68 \\
& FM\cite{FM} + IPv$_{1}+1$ & 53\% & \textbf{38.05} & 74\% & \textbf{1.22} & 55\% & \textbf{2.11} \\
& FM\cite{FM} + IPv$_{1}+2$ & 38\% & \textbf{50.40} & 72\% & \textbf{1.32} & 40\% & \textbf{2.80} \\
& FM\cite{FM} + IPv$_{1}+1+2$ & 52\% & \textbf{39.33} & 79\% & \textbf{1.00} & 47\% & \textbf{2.50} \\
\bottomrule
\end{tabularx}
\end{adjustbox}
\end{table}

\subsubsection{Impact of Modules 1 and 2}
\label{subsub:impact_of_modules_1_and_2}
\begin{figure*}[h]
    \centering
    \includegraphics[width=\textwidth]{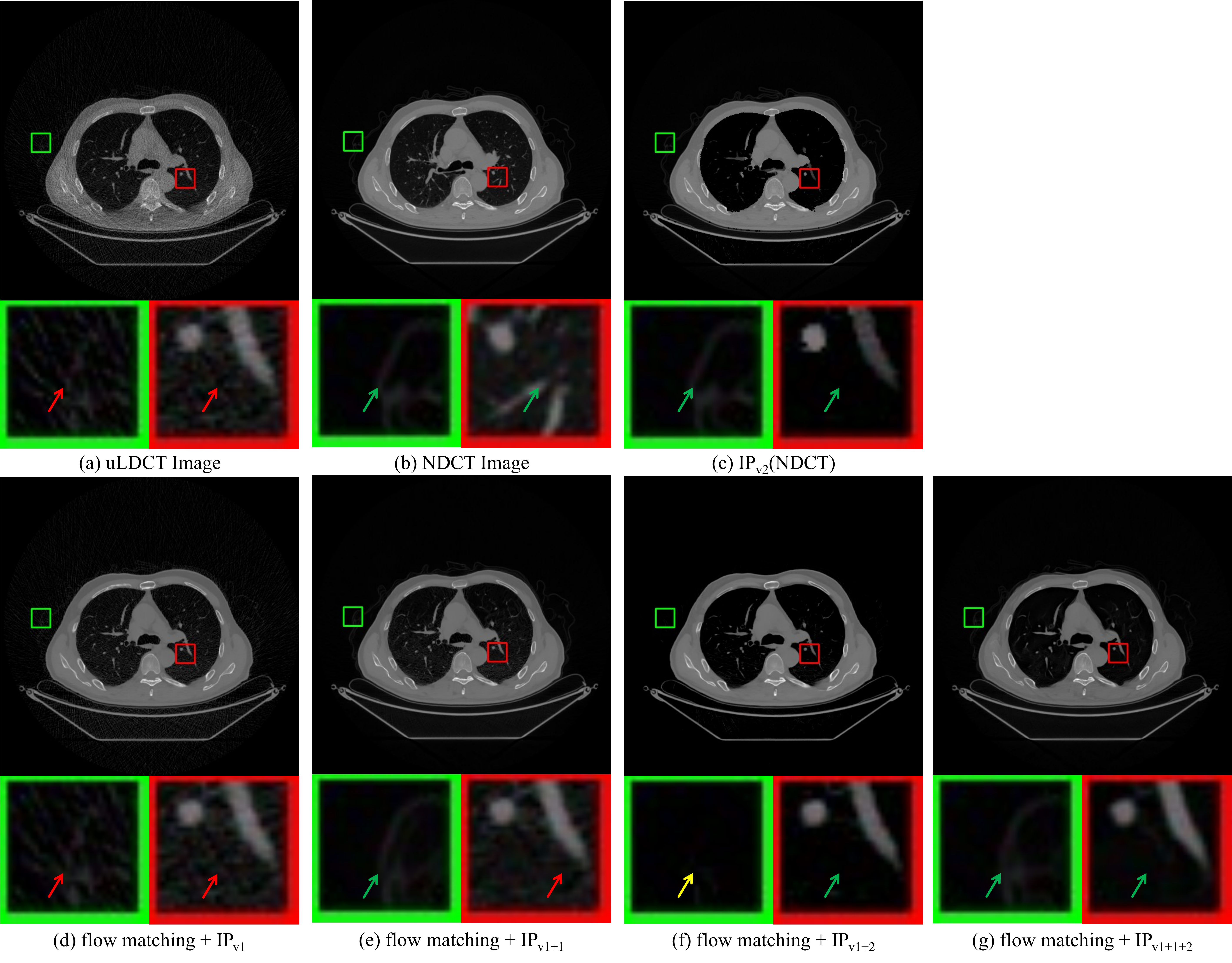}
    \captionsetup{width=\linewidth} 
    \caption{Denoising results on uLDCT using the FM\cite{FM} model with different modules in the image purification strategy. (c) Label constructed with IPv2. (d) to (g) Denoising results of (a) using FM\cite{FM} trained on IPv1 without adding any module, adding Module 1, adding Module 2, and adding both Module 1 and Module 2, respectively. \textcolor{red}{Red} arrows indicate the presence of noise, \textcolor{green}{green} arrows indicate the absence of noise, and \textcolor{yellow}{yellow} arrows indicate the removal of useful information.}
    \label{fig:5}
\end{figure*}

Building on the previous ablation study, we adopt IPv2(NDCT) as the label. As shown in Table~\ref{table:3}, incorporating Module 1 alone leads to an improvement of over 50\% across all perceptual metrics. Adding Module 2 alone yields an improvement exceeding 30\% across all metrics. Combining Module 1 and Module 2 further pushes beyond the performance bottleneck. Figure~\ref{fig:5} illustrates the underlying mechanism. With IPv1 alone, the denoised output contains noise in both the background and lung tissue. Using Module 1 alone effectively removes background noise. Using Module 2 alone removes noise in both regions but at the cost of erasing useful information, a behavior similar to that of Module 3. The key difference is that Module 3 operates on the NDCT images in the test set, whereas Module 2 acts on the uLDCT images in the training set. The joint use of Module 1 and Module 2 removes noise from both the background and lung tissue while preserving useful background information.

\section{CONCLUSION}
\label{sec:CONCLUSION}

This paper addresses the limitation of existing image purification strategy for real-world ultra-low-dose CT denoising, which often fail to adequately denoise both background and lung parenchyma regions. We propose IPv2, a systematically improved strategy that builds upon the core principles of its predecessor by introducing three functional modules: Remove Background, Add Noise, and Remove Noise. These modules collectively equip the model with the capability to perform effective denoising in both background and lung parenchyma areas.

\clearpage

\end{document}